\documentclass[sigconf]{acmart}

\usepackage{booktabs} 
\usepackage{array}
\newcolumntype{P}[1]{>{\centering\arraybackslash}p{#1}}

\usepackage[bottom]{footmisc}
\usepackage{multirow}

\acmArticle{4}
\acmPrice{15.00}

\setcopyright{rightsretained}

\begin{document}
	\title[Adapting Deep Learning for Sentiment Classification]{Adapting Deep Learning for Sentiment Classification of Code-Switched Informal Short Text}

	\author{Muhammad Haroon Shakeel}
	\orcid{0000-0001-6237-3388}
	\affiliation{%
		\institution{Lahore University of Management Sciences}
		\streetaddress{P.O. Box 1212}
		\city{Lahore} 
		\state{Pakistan} 
		\postcode{54792}
	}
	\email{m.shakeel@lums.edu.pk}
	
	\author{Asim Karim}
	\orcid{0000-0002-9872-5020}
	\affiliation{%
		\institution{Lahore University of Management Sciences}
		\streetaddress{P.O. Box 1212}
		\city{Lahore} 
		\state{Pakistan} 
		\postcode{54792}
	}
	\email{akarim@lums.edu.pk}

	\begin{abstract} 
		Nowadays, an abundance of short text is being generated that uses nonstandard writing styles influenced by regional languages. Such informal and code-switched content are under-resourced in terms of labeled datasets and language models even for popular tasks like sentiment classification. In this work, we (1) present a labeled dataset called \textit{MultiSenti} for sentiment classification of code-switched informal short text, (2) explore the feasibility of adapting resources from a resource-rich language for an informal one, and (3) propose a deep learning-based model for sentiment classification of code-switched informal short text. We aim to achieve this without any lexical normalization, language translation, or code-switching indication. The performance of the proposed models is compared with three existing multilingual sentiment classification models. The results show that the proposed model performs better in general and adapting character-based embeddings yield equivalent performance while being computationally more efficient than training word-based domain-specific embeddings.
	\end{abstract}

	\ccsdesc[500]{Computing methodologies~Natural language processing}
	\ccsdesc[300]{Computing methodologies~Language resources}
	
	\copyrightyear{2020}
	\acmYear{2020}
	\acmConference[preprint]{}{January 4}{2020}
	\acmBooktitle{preprint}\acmDOI{xx.xxxx/xxxxxx.xxxxxx}
	\acmISBN{978-1-4503-6866-7/20/03}

	
	\keywords{Deep learning, sentiment classification, code-switching, Roman Urdu, informal language}

	\maketitle
	
	\section{Introduction}
	
	Social media platforms have become popular for sharing sentiments towards a variety of topics. However, the texts on such platforms are often influenced by regional languages. This gives birth to a distinct multilingual phraseology that utilizes informal diction, non-standard abbreviations, improper grammar, and tends to switch between languages mid-utterance, a phenomenon known as code-switching~\cite{fatima2018multilingual}. As a consequence, the task of automatic sentiment classification becomes highly challenging.
	
	Deep learning models have been successful for many NLP tasks involving multilingual and code-switched text. One way to improve the predictive performance of a model is to annotate each word with its respective language (code-switching indications)~\cite{wang-etal-2016-bilingual}. A serious limitation with this approach is its scalability for large data, as the annotation task becomes laborious. More recent approaches translate the under-resourced language into English and then use the resources of the English language to solve the problem in hand~\cite{wang2015emotion,zhou2016attention, chen2018adversarial}. However, this approach is only practical for languages with robust translation resources. Therefore, such strategy is unfeasible for an informal language. It is established that pre-trained word embeddings give a boost to predictive performance of language models~\cite{shakeel2019multicascaded}. However, such \textit{``word-based"} embeddings are limited to English language only with no equivalence for informal languages. An alternative, therefore, is to use \textit{``character-based"} embedding~\cite{arora2019character}. Such embeddings are available in the form of pre-trained models on large scale data of English language, hence are well-suited for any language that uses English alphabets.
	
	The focus of this paper is a specific informal and multilingual dialect of communication known as \textit{``Roman Urdu"}, which utilizes English alphabets to write Urdu and tends to code-switch between English and Urdu. Despite its prevalence, Roman Urdu has received little attention and research on this \textit{``language"} lags behind due to the non-availability of gold-standard datasets.
	
		\begin{table}[!b]
		\caption{MultiSenti dataset characteristics}
		\label{tab:datasetStats}
		\begin{tabular} {cccc} 
			\toprule
			\textbf{Class Label} & \textbf{Class \%age} & \textbf{Language} & \textbf{Language \%age} \\
			\cmidrule(lr){1-2} \cmidrule(lr){3-4}
			Negative &  $48.27\%$ & Roman Urdu & $46.34\%$ \\
			Positive &  $35.10\%$  & English & $2.52\%$ \\
			Neutral &  $16.63\%$ & Mixed & $51.14\%$ \\
			\bottomrule
		\end{tabular}
	\end{table}

	Our first contribution is that we develop an annotated dataset called \emph{MultiSenti} for the problem of sentiment classification of Roman Urdu short text. Our second contribution is that we investigate the feasibility of adapting character-based pre-trained embedding models for sentiment classification of Roman Urdu short text. To exhibit the contrast with adapted embeddings, we also train our own word-based multilingual embeddings on the Roman Urdu corpus. Our third contribution is that we propose a deep learning model for sentiment classification of Roman Urdu short text, namely \emph{McM}. The model tends to learn from raw text only without utilizing lexical normalization, language translation, language transliteration, or code-switching indication. The performance of the proposed model is compared with three existing multilingual sentiment classification models. The results demonstrate that McM outperforms other models in all of the experiments. The study also proves the practicality and usefulness of adapting character-based pre-trained embeddings from English language for Roman Urdu language.

	\section{MultiSenti Dataset} \label{sec:Dataset}
	The MultiSenti dataset is collected from Twitter during and after the general elections of Pakistan in the year $2018$ to identify the overall emotion and sentiment of populous towards the on-going election process and its result. The dataset has been categorized into \textit{``negative"}, \textit{``positive"}, and \textit{``neutral"} sentiments. A sentiment in a tweet can either be expressed in monolingual or multilingual form, i.e., (i) \textit{Roman Urdu}, (ii) \textit{English}, and (iii) \textit{Mixed}. Preprocessing of the data is kept minimal to the extent of lowering the cases and removing all the records having only single word in tweet. The ``gold standard" is constructed by manually annotating $20,735$ samples into predefined categories by two annotators in supervision of a domain-expert. In case of conflict between annotators, decision of domain expert is considered. Class labels percentages and language ratios in the dataset are presented in the Table~\ref{tab:datasetStats}. Class-based stratified sampling at $80-20\%$ is adopted for generating train and test splits of the data. These splits are made available publicly\footnote{https://github.com/haroonshakeel/multisenti}.
	
	\section{Methodology}
	\subsection{Language Resource Adaptation} \label{sec:resource_adaptation}

	\begin{figure*}[!htbp]
		\centering
		\includegraphics[page=1, scale=0.50,clip]{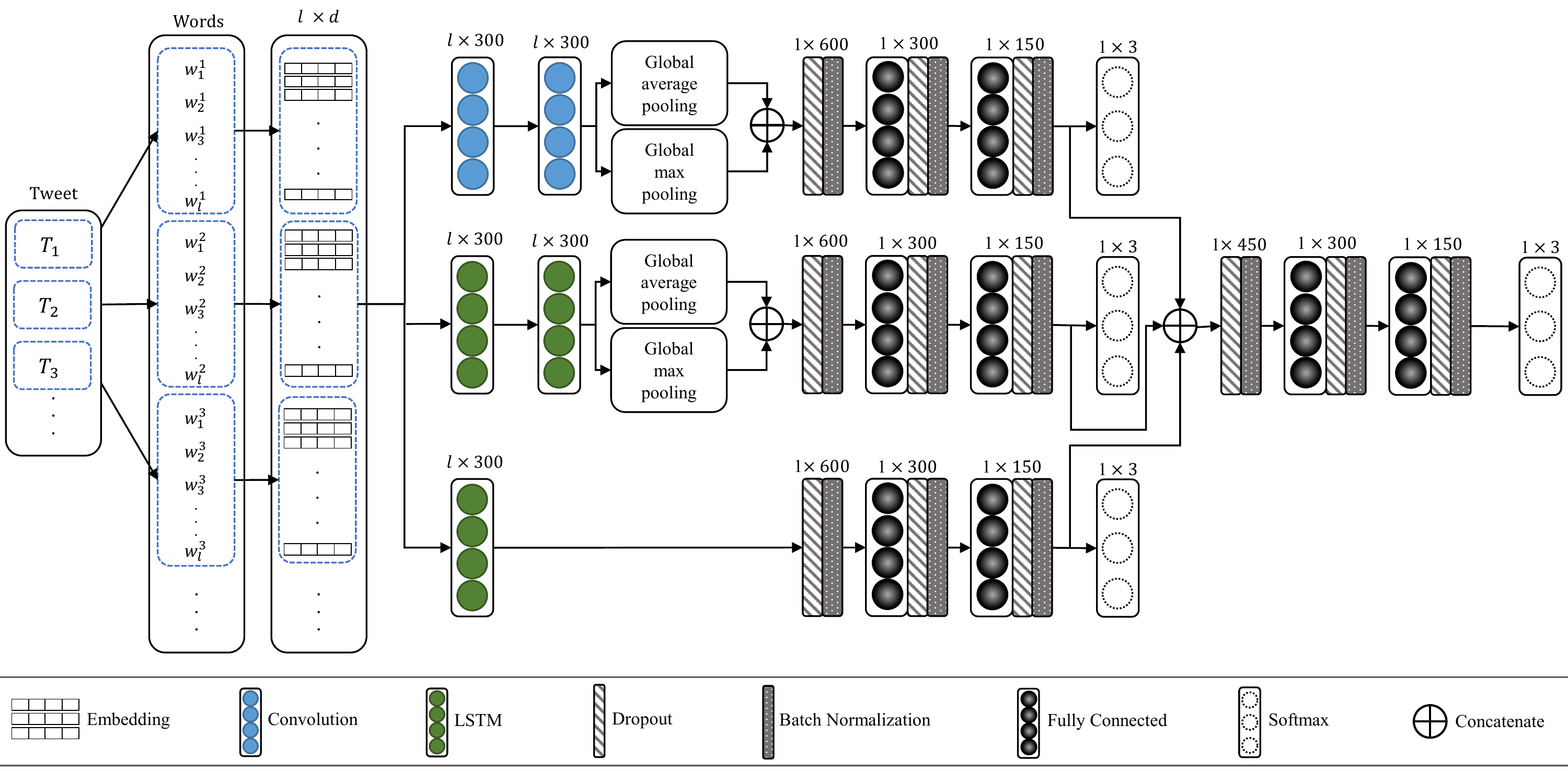}
		\caption{Multi-cascaded model (McM) for sentiment classification of informal short text}
		\label{fig:architectureMcM}
	\end{figure*}
	
	We first examine the feasibility of resource adaptation involving deep learning models and word embedding choices. We select three models with strong predictive performance on multilingual sentiment classification: (i) ConvNets~\cite{medrouk2017deep}, (ii) Attention-LSTM~\cite{zhou2016attention}, and (iii) SimpleConv~\cite{attia2018multilingual}. All models are reimplemented using hyperparameters as defined in the original studies. 
	
	As regards to word embedding choices, a well-known problem is that of ``out-of-vocabulary" where certain words are not found in the embedding base. In such cases, either random initialization of embeddings~\cite{attia2018multilingual} or using character-based pre-trained embeddings~\cite{arora2019character} is plausible. We investigate both strategies on MultiSenti dataset using all three adapted deep learning models mentioned above. For our experiments, random embeddings are initialized from a uniform distribution with $300$ dimensions. The choice of character-based pre-trained embeddings is restricted to ELMo~\cite{peters-etal-2018-deep}, which is trained on a large-scale English language corpus and produces an embedding of $1024$ dimensions. During the training of a model, embedding layer can be finetuned or training can proceed without finetuning~\cite{zhou2016attention}. To assess the out-of-the-box performance of pre-trained embedding model, we also take prediction directly from ELMo by introducing a \textit{softmax} layer on top of it. In this way, a total of $14$ experiments are performed and their results are shown in Table~\ref{tab:results} (lines 1-2 for each adapted model).
	
	The experiments reveal that ELMo out-of-the-box performs on par with the other variations, though finetuning does not affect its predictive performance. However, random embedding initialization benefits from finetuning on all three models. Slightly superior results are achieved when ConvNet and Attention-LSTM models are used on top of ELMo. Interestingly, SimpleConv model shows significant decline in performance when ELMo embedding without finetuning is used. However, with finetuning, it is able to achieve comparable results with other variants. It is also worth noting that employing random embedding without finetuning yields lowest performance. These planned comparisons reveal that finetuning the embedding layer is more beneficial as compared to freezing the weight updates during the training, and using a deep model on top of an embedding layer is a perceptive choice. However, all models on this informal language dataset underperform relative to the results reported on formal languages such as English, French, Greek, and Chinese~\cite{medrouk2017deep, zhou2016attention}. These observations clearly indicate that existing models for formal languages are not well-suited for informal language and call for novel model architectures specifically tailored for informal and code-switched language.

	\subsection{Proposed Model} \label{sec:proposedModel}
	Our proposed deep learning model, called McM, employs three feature learners (cascades) that are trained for classification independently (in parallel) as shown in Figure~\ref{fig:architectureMcM}. The learned features from these learner are forwarded to a discriminator network for final prediction. Each of these four components is discussed below.
	\subsubsection{Stacked-CNN Learner:}
	This learner is employed to learn $n$-gram features for identification of relationships between words. A $1$-d convolution filter is used with a sliding window (kernel) of size $k$ (number of $n$-grams) in order to extract the features. Two CNN layers are stacked which use $k=1$ and $k=2$ respectively. An activation function \textit{ReLU}, which is defined as $\sigma(x) = max(x,0)$, is used to introduce non-linearity. We use $300$ filters and stride $= 1$ for both layers. The output of second CNN layer is followed by (i) global max-pooling to remove low activation information from feature maps of all filters, and (ii) global average-pooling to get average activation across all the $n$-grams. These two outputs are then concatenated and forwarded to a small feedforward network having two fully-connected layers, followed by a \textit{softmax} layer for the prediction of this particular learner. Dropout layer with a rate of $0.5$ and batch-normalization layer is repeatedly used between both fully-connected layers to avoid over-fitting.
	\subsubsection{Stacked-LSTM Learner:}
	 LSTM captures the order information of words where each word is treated as one time step and is fed to LSTM in a sequential manner. While processing the input at the current time step $X_t$, LSTM also takes into account the previous hidden state $h_{t-1}$.  Stacked-LSTM learner is comprised of two LSTM layers. The output of the first LSTM layer is fed to the second LSTM layer and the output produced by second LSTM layer is forwarded to global max-pooling and global-average pooling layers. The former drops the low activations while the latter averages activations across all time steps. These two outputs are concatenated and forwarded to a two-layered feedforward network for intermediate supervision, identical to previously described stacked-CNN learner. We use $300$ LSTM units in both layers.
	 \subsubsection{LSTM learner:}
	This learner is employed to learn long-term dependencies of the text as described in~\cite{wang-etal-2016-combination}. This learner encodes complete input text recursively and returns a single vector. The dimensions of the output vector are equal to the number of LSTM units deployed. This encoded text representation is then forwarded to a small feedforward network identical to the aforementioned two learners, for intermediate supervision in order to learn features. This learner differs from stacked-LSTM learner as it learns sentence features, not average and max-features of all time steps (input words). This learner uses $300$ LSTM units.
	\subsubsection{Discriminator Network:}
	This small feedforward network aggregates features learned by each of the above described three learners and squash them into a small network for final prediction. It employs two fully-connected layers with dropout and batch-normalization layer along with \textit{ReLU} activation function for non-linearity. The \textit{softmax} activation function with categorical cross-entropy loss is used on the final prediction layer to get probabilities of each class. The class label is assigned based on maximum probability. This is treated as the final prediction of the proposed model. Note that the choices of the number of convolutional filters, number of units in dense layers, and number of LSTM units are made empirically. Rest of the hyperparameters (choices of $k$, dropout rate, optimizer, and learning rate) were selected by performing a grid search using a $20\%$ stratified validation set taken from training set and utilizing random embedding initialization without finetuning. The complete architecture, along with dimensions of each output is shown in Figure~\ref{fig:architectureMcM}. The network is optimized using ``Adam" optimizer with a learning rate of $0.002$.

	\begin{table*}[!tp]
		\centering
		\setlength{\tabcolsep}{3pt}
		\caption{Performance evaluation of variations of the proposed models and baselines. (Showing highest scores in boldface.)}
		\label{tab:results}
		\begin{tabular}{llcccccccc}
			
			\toprule
			& & \multicolumn{4}{c}{\textbf{Without Finetuning}} & \multicolumn{4}{c}{\textbf{With Finetuning}} \\
			\cmidrule(r){3-6} \cmidrule(l){7-10}
			{\bfseries Model} & {\bfseries Embedding} & {\bfseries Accuracy} & {\bfseries Precision} & {\bfseries Recall} & {\bfseries F1-score} & {\bfseries Accuracy} & {\bfseries Precision} & {\bfseries Recall} & {\bfseries F1-score} \\
			
			\toprule
			ELMo~\cite{peters-etal-2018-deep} & $-$ & $0.63$ & $0.66$ & $0.55$ & $0.57$ & $0.63$ & $0.64$ & $0.55$ & $0.57$ \\
			\midrule
			\multirow{3}{*}{ConvNet~\cite{medrouk2017deep}} & Random & $0.60$ & $0.57$ & $0.58$ & $0.57$ & $0.63$ & $0.60$ & $0.61$ & $0.61$ \\
			& ELMo & $0.64$ & $0.63$ & $0.59$ & $0.60$ & $0.64$ & $0.62$ & $0.60$ & $0.61$ \\
			& Multilingual & $0.63$ & $0.60$ & $0.57$ & $0.58$ & $0.65$ & $0.62$ & $0.60$ & $0.61$ \\
			
			\midrule
			\multirow{3}{*}{Attention-LSTM~\cite{zhou2016attention}} & Random & $0.59$ & $0.56$ & $0.56$ & $0.55$ & $0.66$ & $0.64$ & $0.64$ & $0.64$ \\
			& ELMo & $0.66$ & $0.64$ & $0.61$ & $0.62$ & $0.64$ & $0.63$ & $0.60$ & $0.61$ \\
			& Multilingual & $0.64$ & $0.61$ & $0.61$ & $0.61$ & $0.66$ & $0.63$ & $0.62$ & $0.62$ \\

			\midrule
			\multirow{3}{*}{SimpleConv~\cite{attia2018multilingual}} & Random & $0.64$ & $0.63$ & $0.58$ & $0.60$ & $0.67$ & $0.65$ & $0.63$ & $0.63$ \\
			& ELMo & $0.17$ & $0.06$ & $0.33$ & $0.10$ & $0.62$ & $0.65$ & $0.54$ & $0.55$ \\
			& Multilingual & $0.35$ & $0.12$ & $0.33$ & $0.17$ & $0.35$ & $0.12$ & $0.33$ & $0.17$ \\
			
			\midrule
			\multirow{3}{*}{McM} & Random & ${0.67}$ & ${\textbf{0.67}}$ & ${0.61}$ & ${0.62}$ & $0.67$ & $0.69$ & $0.60$ & $0.62$ \\
			& ELMo &${0.67}$ & ${0.66}$ & ${0.62}$ & ${0.63}$ & $0.68$ & $0.65$ & $\textbf{0.65}$ & $\textbf{0.65}$ \\
			& Multilingual &${\textbf{0.68}}$ & ${\textbf{0.67}}$ & ${\textbf{0.64}}$ & ${\textbf{0.65}}$ & $\textbf{0.69}$ & $\textbf{0.71}$ & $0.62$ & $0.64$ \\

			\bottomrule
		\end{tabular}
	\end{table*}

	\subsection{Multilingual Embeddings}\label{subsec:multilinguan_embeddings}
	
	We also compare multilingual embeddings constructed from a combined corpus constituted of MultiSenti dataset and another large scale Roman Urdu dataset (These embeddings are made available along with dataset). The total number of words in this combined corpus was more than $6.5$ million. We use skip-gram model of word2vec with word vector of size $d = 300$ as suggested in original study~\cite{mikolov2013distributed}. These embeddings are trained for $500,000$ iterations.

	\subsection{Implementation Details}\label{subsec:evaluation_metrics}
	
	All the implementation is done in Python using Keras library with Tensorflow backend. All weights of the networks are initialized randomly and to mitigate the effect of randomness, random seed is fixed across all experiments. For every experiment, the model is trained for $100$ epochs. A checkpoint of the learned weights is saved at epoch with the best predictive performance on the test split. The early stopping approach is also opted and training is stopped if testing error does not decrease for $10$ epochs.
	
	\section{Results and Conclusion} \label{sec:Results}
	
	We report performance of all variations using \textit{accuracy} and macro-averaged  \textit{precision, recall,}, and \textit{F1-score} in Table~\ref{tab:results}. In the discussion, however, we focus on F1-score . Based on the results, we make the following observations.
	
	Using pre-trained embeddings out-of-the-box yields identical performance when used either without or with finetuning. Specifically talking about the case when a model is used on top of the embeddings, ELMo embeddings without finetuning outperform the random and multilingual embeddings on ConvNet and Attention-LSTM. Interestingly, in the case of SimpleConv model, ELMo yields the poorest performance. Further examination of this particular case revealed that this particular model is unable to learn when pre-trained embeddings are used. However, using random embeddings for SimpleConv gives output comparable to other models. This implies that the model is specifically engineered to work with random embedding (as is evident from the original study). Regarding the use of random embedding for other models, the proposed model McM achieves highest F1-score. Amongst rest of the models, ConvNet marginally outperforms Attention-LSTM. As far as the use of multilingual embeddings is concerned, it was found that the least F1-score was achieved by SimpleConv, while McM achieved the highest score, which surpasses all the experiments without finetuning the embeddings.
	
	Turning now to the case of finetuning, ConvNet performs identical in terms of F1-score for all of the embeddings, while Attention-LSTM and SimpleConv benefit from finetuning when random embeddings are used. In regards to McM model, ELMo embeddings yield the highest F1-score of $0.65$. This is an interesting finding as it is identical to the F1-score of McM when multilingual embedding without finetuning is used. 
	
	It is worth noting that even though simpler networks such as ConvNets and SimpleConv take the least amount of training time, their performance is inconsistent across all settings. While the proposed model McM shows the highest performance in the majority of the cases with $\pm 3\%$ variation for each embedding. These findings lead to conclude that no apparent advantage exists in training word-based multilingual embeddings from scratch. The pre-trained character-based embedding on the English language with finetuning suffices for informal language to get identical results while avoiding pre-training overhead. However, to get most out of these embedding, a carefully tailored model for sentiment classification of informal short text is crucial. One can argue that embeddings trained on an informal multilingual corpus, which is comparable in size to the corpus of English language, could yield better performance than adapting the embeddings. However, this leads to the initial paradox of not having enough data resources for the informal languages.

	Our work has led us to conclude that adapting existing resources from a resource-rich language to an informal language is practical. It is evident from the results that an embedding trained on sufficiently large corpus in the English language can successfully be adapted for an informal language. However, this is not necessarily true for the model choice. It is crucial that a model is engineered specifically towards an informal language as compared to adapting models developed for other languages. As future research, we plan to investigate other embedding choices such as BERT~\cite{devlin2019bert}.

	\bibliographystyle{ACM-Reference-Format}
	\bibliography{mybibliography} 
	
\end{document}